\documentclass{article}

\PassOptionsToPackage{numbers}{natbib}

\usepackage[final]{nips_2018}
\usepackage{epsfig}
\usepackage{graphicx,amsmath,amssymb,amsthm,xspace,microtype,soul}
\usepackage[utf8]{inputenc} 
\usepackage[T1]{fontenc}    
\usepackage{hyperref}       
\usepackage{url}            
\usepackage{booktabs}       
\usepackage{amsfonts}       
\usepackage{nicefrac}       
\usepackage{microtype}      

\title{Learning the progression and clinical subtypes of Alzheimer’s disease from longitudinal clinical data}

\author{
  Vipul Satone$^1$\\
   \And
  Rachneet Kaur$^1$\\
    \And
  Faraz Faghri$^{1,2}$\\
    \And
  Mike A Nalls$^2$\\ 
   \And
   Andrew B Singleton$^2$\\ 
   \And  
  Roy H Campbell$^1$\\
    \And
  for the Alzheimer’s Disease Neuroimaging Initiative\thanks{Data used in preparation of this article were obtained from the Alzheimer’s Disease Neuroimaging Initiative (ADNI) database (adni.loni.usc.edu). As such, the investigators within the ADNI contributed to the design and implementation of ADNI and/or provided data but did not participate in analysis or writing of this report. A complete listing of ADNI investigators can be found at: http://adni.loni.usc.edu/wp-content/uploads/how\_to\_apply/ADNI\_Acknowledgement\_List.pdf}\\ 
  \\
  \and
  $^1$ University of Illinois at Urbana-Champaign, USA\\
 $^2$ National Institutes of Health, USA\\
}

\begin{document}
\maketitle
\begin{abstract}
Alzheimer’s disease (AD) is a degenerative brain disease impairing a person’s ability to perform day to day activities. The clinical manifestations of Alzheimer’s disease are characterized by heterogeneity in age, disease span, progression rate, impairment of memory and cognitive abilities. Due to these variabilities, personalized care and treatment planning, as well as patient counseling about their individual progression is limited.  Recent developments in machine learning to detect hidden patterns in complex, multi-dimensional datasets provides significant opportunities to address this critical need. In this work, we use unsupervised and supervised machine learning approaches for subtype identification and prediction. We apply machine learning methods to the extensive clinical observations available at the Alzheimer’s Disease Neuroimaging Initiative (ADNI) data set to identify patient subtypes and to predict disease progression. Our analysis depicts the progression space for the Alzheimer’s disease into low, moderate and high disease progression zones. The proposed work will enable early detection and characterization of distinct disease subtypes based on clinical heterogeneity. We anticipate that our models will enable patient counseling, clinical trial design, and ultimately individualized clinical care. 
\end{abstract}

\section{Introduction}
Alzheimer’s disease (AD) is an irreversible and age-associated neurodegenerative disease. Alzheimer’s disease is the most common form of dementia which progressively affects memory, intellectual skills, and other mental functions. Researchers have shown that AD initiates 10-15 years before the actual disease symptoms are manifested \cite{allaire1999everyday}. In the absence of any cure, early detection of AD onset and progression rate is significant as it helps in designing disease modifying treatment strategies. A major challenge for AD prediction and individualized clinical care is the phenotypic heterogeneity that exists within the AD population. 

Attempts thus far at the characterization of disease subtypes have followed a path of clinical observation based cognitive abilities. The disease is often separated into dementia and mild cognitive impairment (MCI), a stage between expected cognitive decline of normal aging and a more serious decline of dementia. MCI patients are at a higher risk of progressing to dementia, but not all the patients end up developing AD \cite{hassan2017machine}. Previous studies attempted to detect incipient AD in patients with MCI or predict early stage of the AD using Cerebrospinal Fluid (CSF) \cite{hansson2006association}, \cite{hassan2017machine}. Psychometric and imaging data has also been used for predicting the progression of dementia in patients with amnestic mild cognitive impairment \cite{Moreland}. However, less work has been done on using just clinical data and predicting the Alzheimer’s disease progression rate. 

We have previously used multimodal clinical data to produce progression space for Parkinson’s Disease \cite{faghri2018predicting}. Here we describe our work on clustering and early prediction of the clinical progression of the AD. We have used dimension reduction techniques to identify progression space, Gaussian Mixture Model (GMM) for subtyping, and Random Forest to predict an individual’s progression. This work has identified Alzheimer’s Disease progression space with three clinically distinct disease subtypes corresponding to memory decline and cognitive decline. Following the successful creation of disease subtypes within a progression space, we created a baseline predictor that accurately predicts an individual patient’s clinical group membership two years later.

\section{Methods}
\subsection{Study design and participants}
This study included data from the Alzheimer’s Disease Neuroimaging Initiative (ADNI) database (adni.loni.usc.edu). The ADNI was launched in 2003 as a public-private partnership, led by Principal Investigator Michael W. Weiner, MD. For up-to-date information, see www.adni-info.org. Machine learning models are developed using baseline and 12 months of data, for each patients’ position after 24 and 48 months. The study consist of 248 cases (including 123 (49.59\%) female, average age for all participants is 71.56 $\pm$ 6.78 years, 94.75\% of them are of European ancestry) for prediction at $48^{th}$  month and 453 cases (including 208 (45.91\%) female, average age for all participants is 72.34 $\pm$ 7.12 years, 93.59\% of them are of European ancestry) for prediction at $24^{th}$  month. Cases in $24^{th}$  month have a mean age of 72.89 $\pm$ 6.06, 71.61 $\pm$ 7.49 and 72.92 $\pm$ 8.11 corresponding to controls, MCI, and dementia patients. Cases in $48^{th}$  month have a mean age of 72.18 $\pm$ 6.63, 71.36 $\pm$ 6.67 and 70.34 $\pm$ 7.42 corresponding to control, MCI and dementia patients. We used comprehensive and  longitudinally collected measurements including: Montreal Cognitive Assessment \cite{nasreddine2005montreal}, Clinical Dementia Rating \cite{berg1984clinical}, Neuropsychiatric Inventory Questionnaire \cite{de2003neuropsychiatric}, Neuropsychological Battery \cite{dodrill1978neuropsychological}, Mini Mental State Exam \cite{burns1998mini}, Geriatric depression scale \cite{gd}, Everyday cognition - study partner \cite{salomon1991partners}, Everyday cognition - participant \cite{allaire1999everyday} and Functional Assessment Questionnaire \cite{fillenbaum1981development}.

\subsection{Procedures and statistical analysis}
Cases with data recorded for all the measurements were included. After imputations using domain knowledge, data cleansing and preprocessing, 218 clinical features were retained. If required, one hot encoding was used for all the categorical variables. Min-max normalization was used to retain progression trend in multimodal measurements. To reduce the dimensionality of the data set, Non-negative Matrix Factorization (NMF) \cite{wang2013nonnegative} (with a rank of 2) was used. NMF reduced high dimensional data into a two-dimensional space. The transformed data were used to project the patient’s disease progression stage at the $24^{th}$  and $48^{th}$  month. Coefficient matrix obtained from the NMF was used as a progression indicator to interpret the representation of the reduced space dimensions.

Unsupervised clustering method Gaussian Mixture Model (GMM) \cite{mclachlan1988mixture} was used to identify clusters. Bayesian Information Criterion (BIC) \cite{schwarz1978estimating} was used to select the optimum number of clusters in GMM. After obtaining the progression space and classifying MCI and dementia patients into different progression subtypes, various supervised learning classifiers were compared to predict an individual patient’s progression. We validate the effectiveness of our predictive models using a five-fold cross-validation and measuring the AUC of the ROC curve.

\begin{figure}
\centering
\begin{minipage}{0.33\columnwidth}
    \includegraphics[height = 4.8cm, width =  \columnwidth]{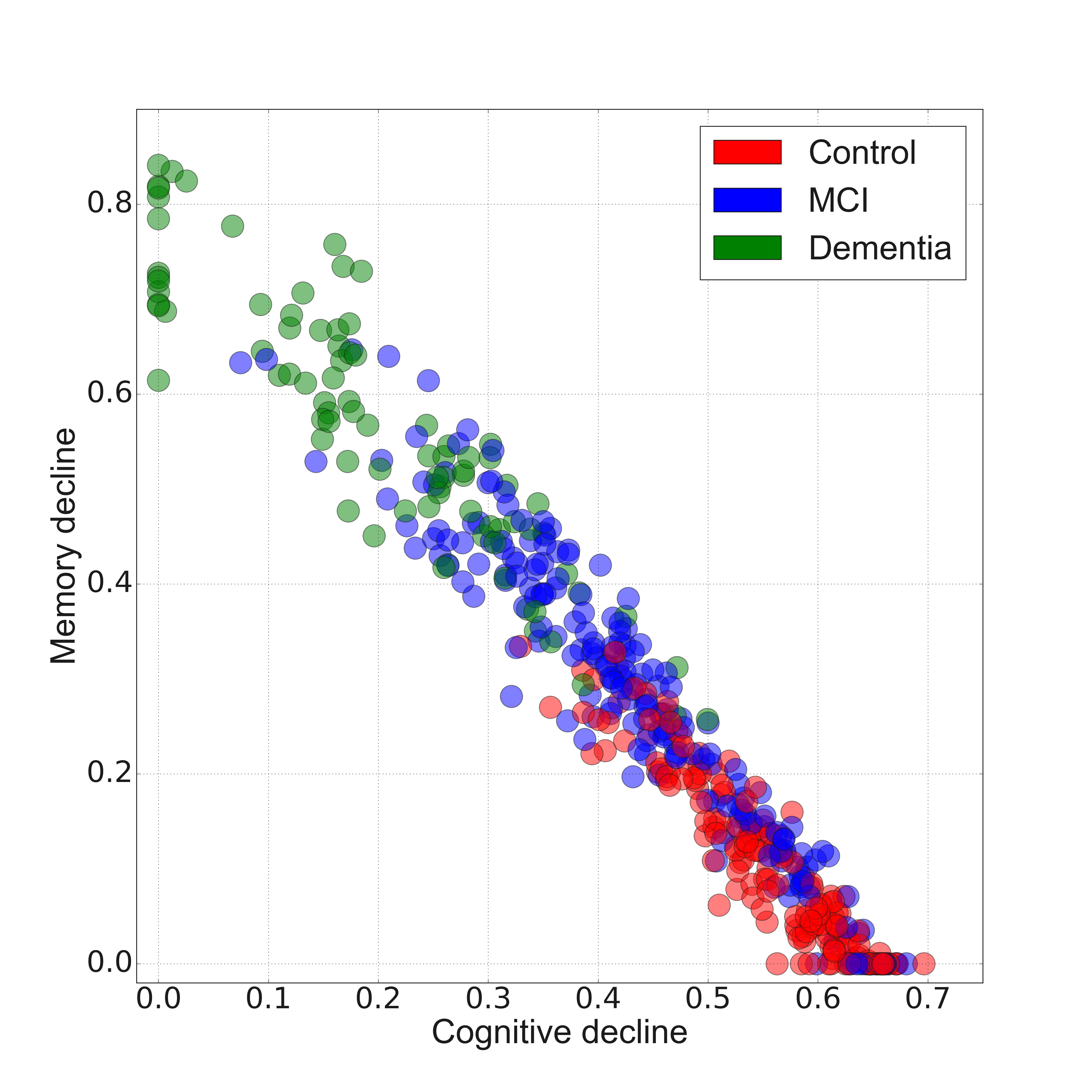}
\end{minipage}
\begin{minipage}{0.37\columnwidth}
    \includegraphics[height = 4.9cm, width = \columnwidth]{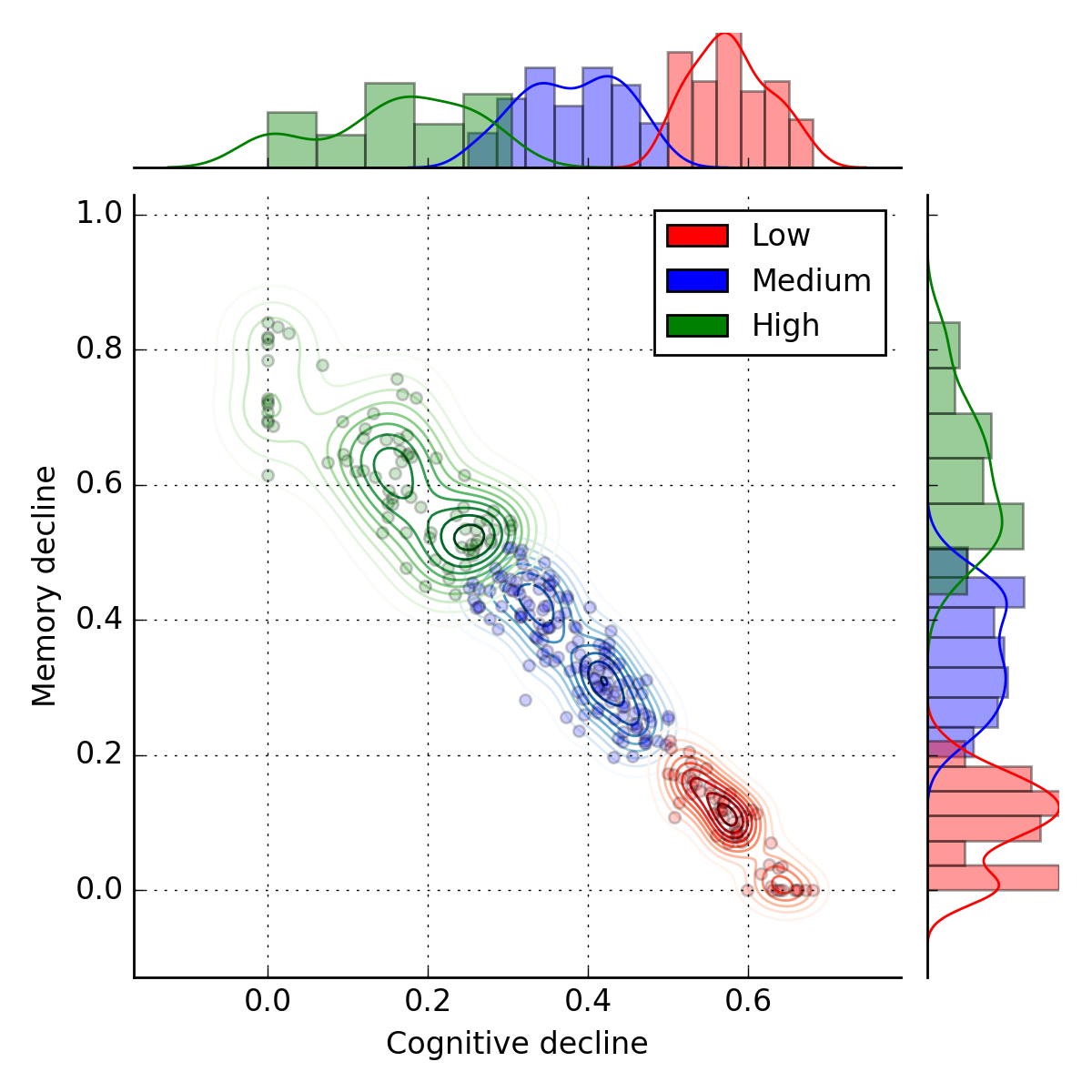}
\end{minipage}
    \caption{\textbf{Left:} Total 453 cases are projected in the Alzheimer’s Disease Progression Space at $24^{th}$  month. Controls are represented in red, MCI in blue and dementia in green.
   \textbf{Right:} Three different progression rates are identified in MCI and dementia patients at $24^{th}$  month. In this figure low progression rate zone is represented in red, moderate progression rate in blue and high progression rate in green.}
    \label{fig:nmf_2}
\end{figure}

\begin{figure}[ht]
\centering
\begin{minipage}{0.35\columnwidth}
    \includegraphics[height = 4.6cm, width =  \columnwidth]{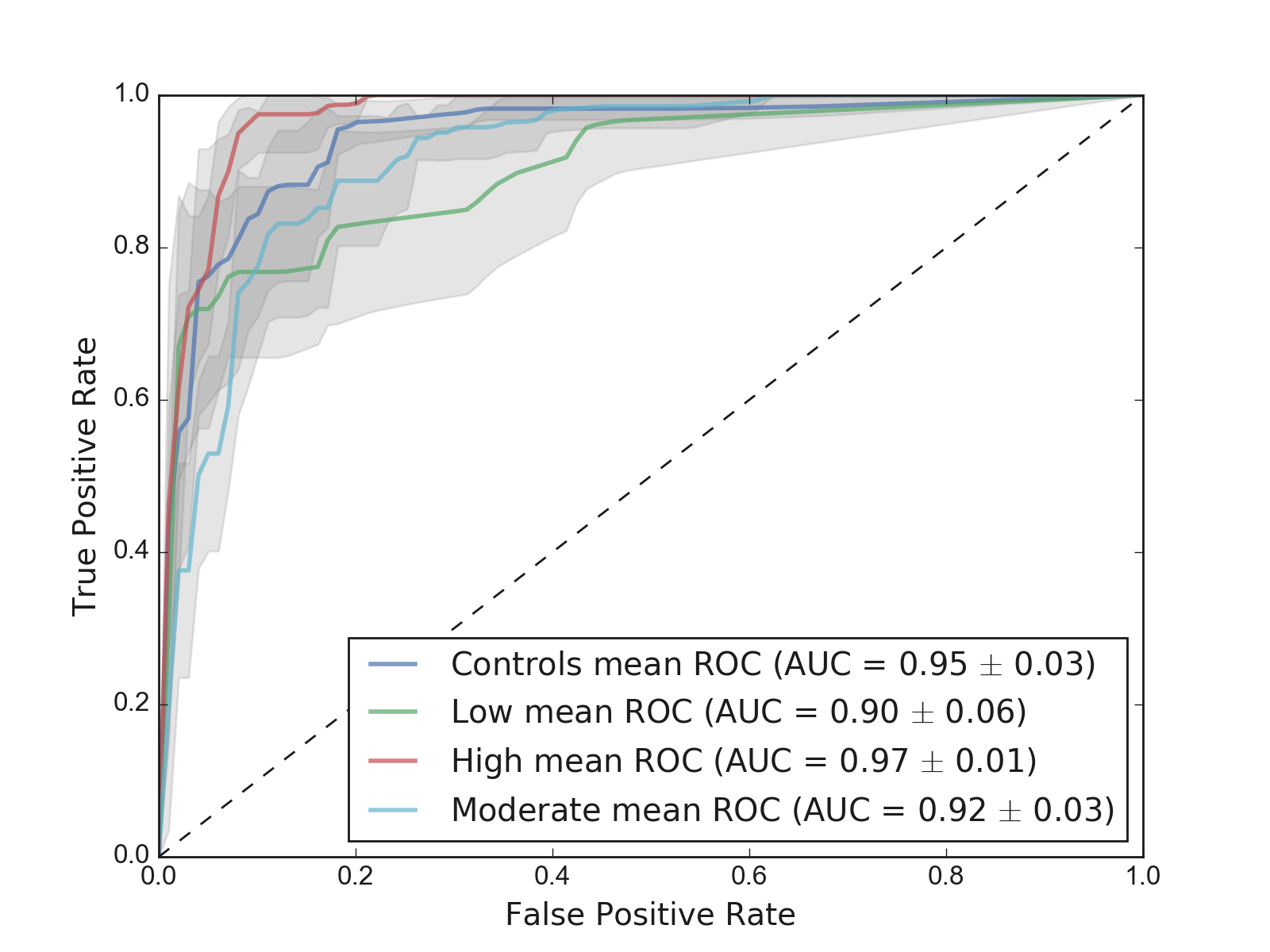}
\end{minipage}
\begin{minipage}{0.37\columnwidth}
    \includegraphics[height = 5.0cm, width = \columnwidth]{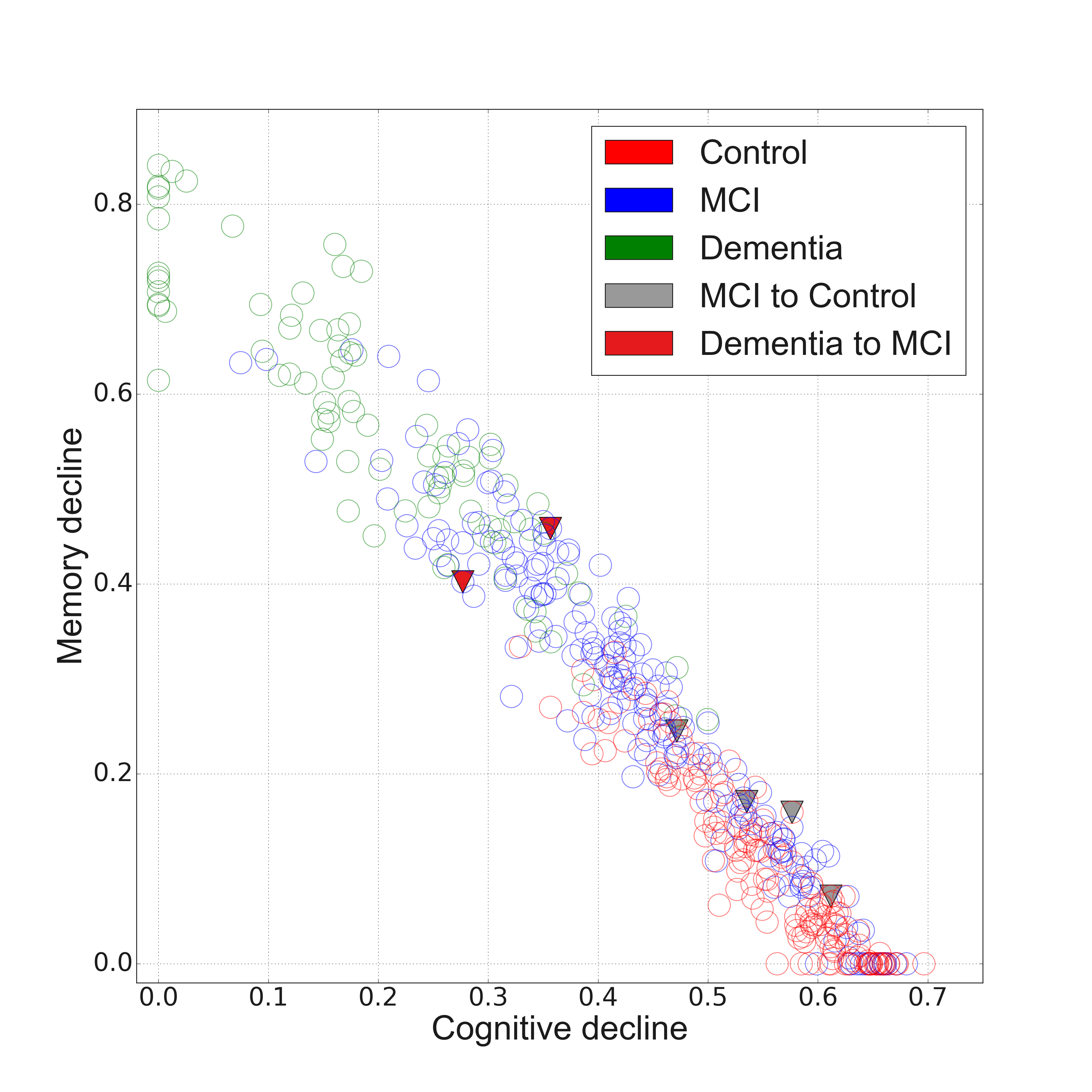}
\end{minipage}
    \caption{\textbf{Left:} Area under ROC curve for each individual class. The predictions for disease stage at $24^{th}$  month were made using Random Forest Algorithm. 
   \textbf{Right:} Shows positions of the reversions (only MCI to control and dementia to MCI) at $24^{th}$  month.}
    \label{fig:roc_reversion}
\end{figure}

\begin{table}[ht]
  \caption{Accuracy of the two models}
  \label{accuracy}
  \centering
  \begin{tabular}{llllll}
    \toprule
    \cmidrule(r){1-2}
 & Accuracy & AUC -control & AUC -low & AUC -moderate & AUC -high
 \\
    \midrule
    M24 & 84.54 $\pm$  3.35  & 0.95 $\pm$ 0.03 &  0.90 $\pm$ 0.06   & 0.92 $\pm$ 0.03 & 0.97 $\pm$ 0.01  \\
    M48     & 80.24 $\pm$ 4.33 & 0.90 $\pm$ 0.04 & 0.84 $\pm$ 0.03  & 0.88 $\pm$ 0.03   & 0.96 $\pm$ 0.03     \\
    \bottomrule
  \end{tabular}
\end{table}

\section{Results \& Discussion}
Figure \ref{fig:nmf_2} demonstrates \textit{Alzheimer's disease progression space} for $24^{th}$  month, generated by the dimension reduction step. Using the values from the coefficient matrix, the latent features for the x and y-axis were interpreted and identified as cognition decline and memory decline representations. Along the positive direction of y-axis, the memory decline increases and along the negative direction of x-axis the cognitive decline increases. Therefore, the progression rate increases in the direction of the positive y-axis and negative x-axis. Using GMM, three optimum clusters corresponding to low, moderate and high progression rate were identified. Figure \ref{fig:nmf_2} shows the distribution of the three clusters in MCI and dementia patients. Random forest (RF) classifier \cite{breiman2001random} provided the most accurate model. RF parameters were fine-tuned using grid search and evaluated by 5 fold cross-validation accuracy. Two models were developed for predicting progression after 24 and 48 months (accuracy results are presented in Table \ref{accuracy}). Figure \ref{fig:roc_reversion} shows the area under ROC curve of the progression prediction model.

ADNI is a longitudinal study in which the disease state of patients is assessed every 12 months. Generally, most patients’ condition either deteriorated or stayed the same. However, in rare cases, the patient’s condition has reversed to a better state. Essentially, some patients were observed moving from dementia to MCI or MCI to control stage. Interestingly, this behavior is been captured in our progression space. Figure \ref{fig:roc_reversion} shows these reversion cases in the progression space. Patients moving from dementia to MCI fall in the intermediate region between dementia and MCI (moderate progression rate region). Similarly, patients moving from MCI to control lie in intermediate progression region between MCI and control. This observation signifies the robustness of our proposed progression space model. 

\begin{figure}
\centering
\begin{minipage}{0.39\columnwidth}
    \includegraphics[height = 5.4cm, width =  \columnwidth]{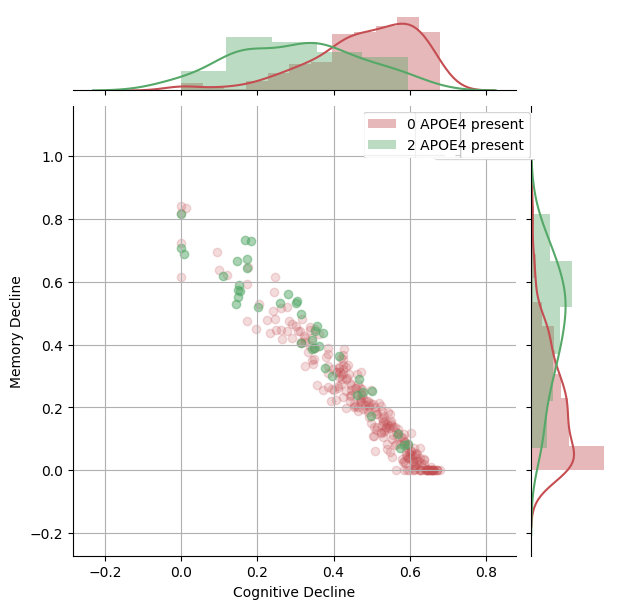}
\end{minipage}
\begin{minipage}{0.37\columnwidth}
    \includegraphics[height = 5.4cm, width = \columnwidth]{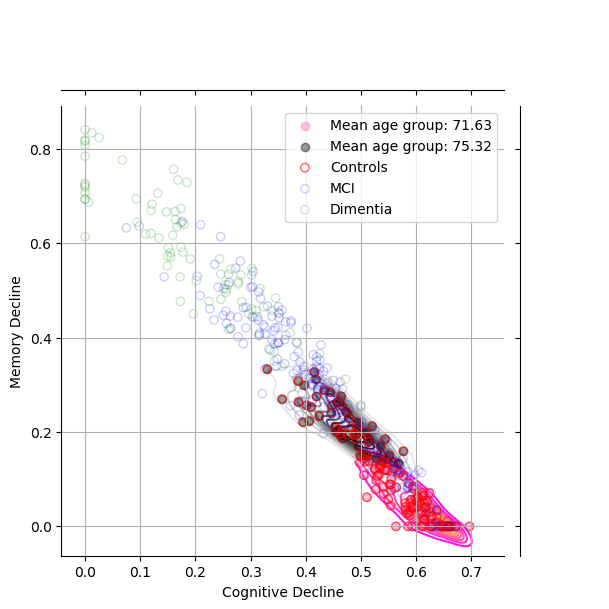}
\end{minipage}
    \caption{\textbf{Left:} Projection of AD cases with the different number of APOE4 variants to the projection space at $24^{th}$  month. 
   \textbf{Right:} Shows the mean age of cases in the two clusters of controls. Clusters represent aging pattern in controls at $24^{th}$  month. The cluster near moderate progression rate zone has high mean age than one which is away from it.}
    \label{fig:apoe_age}
\end{figure}

In another observation, we see the effect of Apolipoprotein E (APOE) genetic mutation. APOE comes in three variants namely, 2, 3, and 4, out of which APOE4 gene variant is closely associated with higher risk of Alzheimer’s disease \cite{strittmatter1993binding}. Figure \ref{fig:apoe_age} shows the cases with 0 and 2 counts of APOE4 variants in our AD progression space. As evident from the figure, observations with 0 count of APOE4 variant are concentrated towards the low progression rate zone, whereas observations with 2 counts of APOE4 variant are concentrated towards moderate and high disease progression rate. This observation is validated by the existing literature which has identified a significant correlation between APOE4 genetic variants and cognitive performance \cite{Raber}. Noting that APOE4 data was not considered during the construction of the projection space as only the time-variant measurements are used to obtain progression space. 

Using our progression space model we can also observe the aging pattern in controls. Figure \ref{fig:apoe_age} shows the normal cognition and memory decline in controls attributed to increase of age. Since this decline is not severe, control cases do not lie in moderate or high progression rate zones. A simple clustering of observations into two clusters shows the stark difference in the mean age of the clusters. It is interesting to note that the mean age for the cluster which is relatively close to the moderate progression rate zone is 75.25 years and mean age of the cluster away from the moderate progression rate zone is 70.75 years.

\section{Conclusions}
Prediction of disease and disease course is a critical challenge in the care, treatment, and research of complex heterogeneous diseases such as Alzheimer's. In this work, we used machine learning techniques to identify Alzheimer's disease subtypes based on distinct progression stages. Furthermore, we developed predictive models of the progression stage after 24 and 48 months from baseline. The developed machine learning solutions potentially offer substantial clinical impact by augmenting clinical decision-making for physicians and healthcare specialists.

\section{Acknowledgement}
This work was supported in part by the Intramural Research Programs of the National Institute on Aging (NIA)Z01-AG000949-02. Data collection and sharing for this project was funded by the Alzheimer's Disease Neuroimaging Initiative (ADNI) (National Institutes of Health Grant U01 AG024904) and DOD ADNI (Department of Defense award number W81XWH-12-2-0012). ADNI is funded by the National Institute on Aging, the National Institute of Biomedical Imaging and Bioengineering, and through generous contributions from the following: AbbVie, Alzheimer’s Association; Alzheimer’s Drug Discovery Foundation; Araclon Biotech; BioClinica, Inc.; Biogen; Bristol-Myers Squibb Company; CereSpir, Inc.; Cogstate; Eisai Inc.; Elan Pharmaceuticals, Inc.; Eli Lilly and Company; EuroImmun; F. Hoffmann-La Roche Ltd and its affiliated company Genentech, Inc.; Fujirebio; GE Healthcare; IXICO Ltd.; Janssen Alzheimer Immunotherapy Research and Development, LLC.; Johnson and Johnson Pharmaceutical Research and Development LLC.; Lumosity; Lundbeck; Merck and Co., Inc.; Meso Scale Diagnostics, LLC.; NeuroRx Research; Neurotrack Technologies; Novartis Pharmaceuticals Corporation; Pfizer Inc.; Piramal Imaging; Servier; Takeda Pharmaceutical Company; and Transition Therapeutics. The Canadian Institutes of Health Research is providing funds to support ADNI clinical sites in Canada. Private sector contributions are facilitated by the Foundation for the National Institutes of Health (www.fnih.org). The grantee organization is the Northern California Institute for Research and Education, and the study is coordinated by the Alzheimer’s Therapeutic Research Institute at the University of Southern California. ADNI data are disseminated by the Laboratory for Neuro Imaging at the University of Southern California.

\bibliography{sample} 
\bibliographystyle{plainnat}
\newpage
\centering
\section{Supplementary Material}

\begin{figure}[ht]
    \includegraphics[height = 4.5cm, width =  15cm]{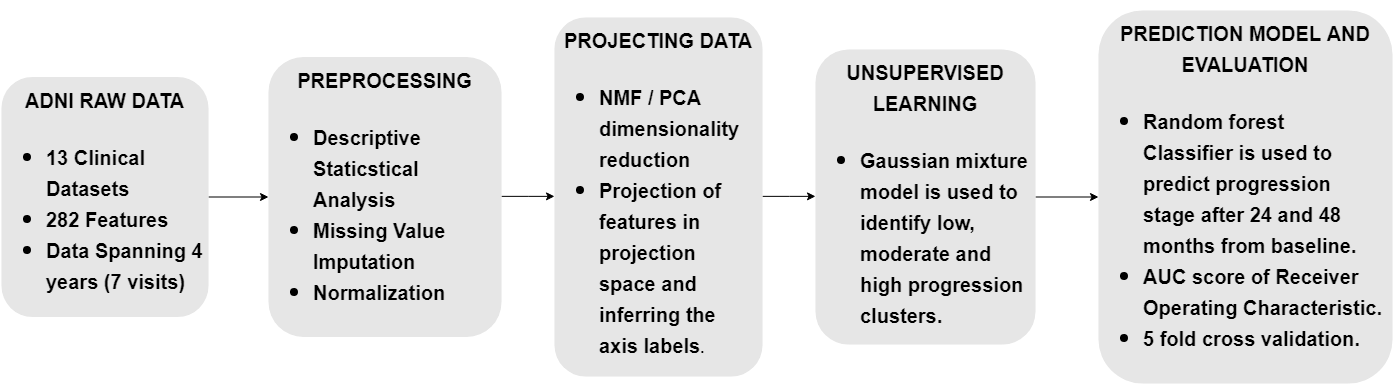}
    \caption{Our machine learning analysis workflow }
    \label{flowchart}
\end{figure}

\begin{figure}[ht]
\begin{minipage}{0.43\columnwidth}
    \includegraphics[height = 6cm, width =  \columnwidth]{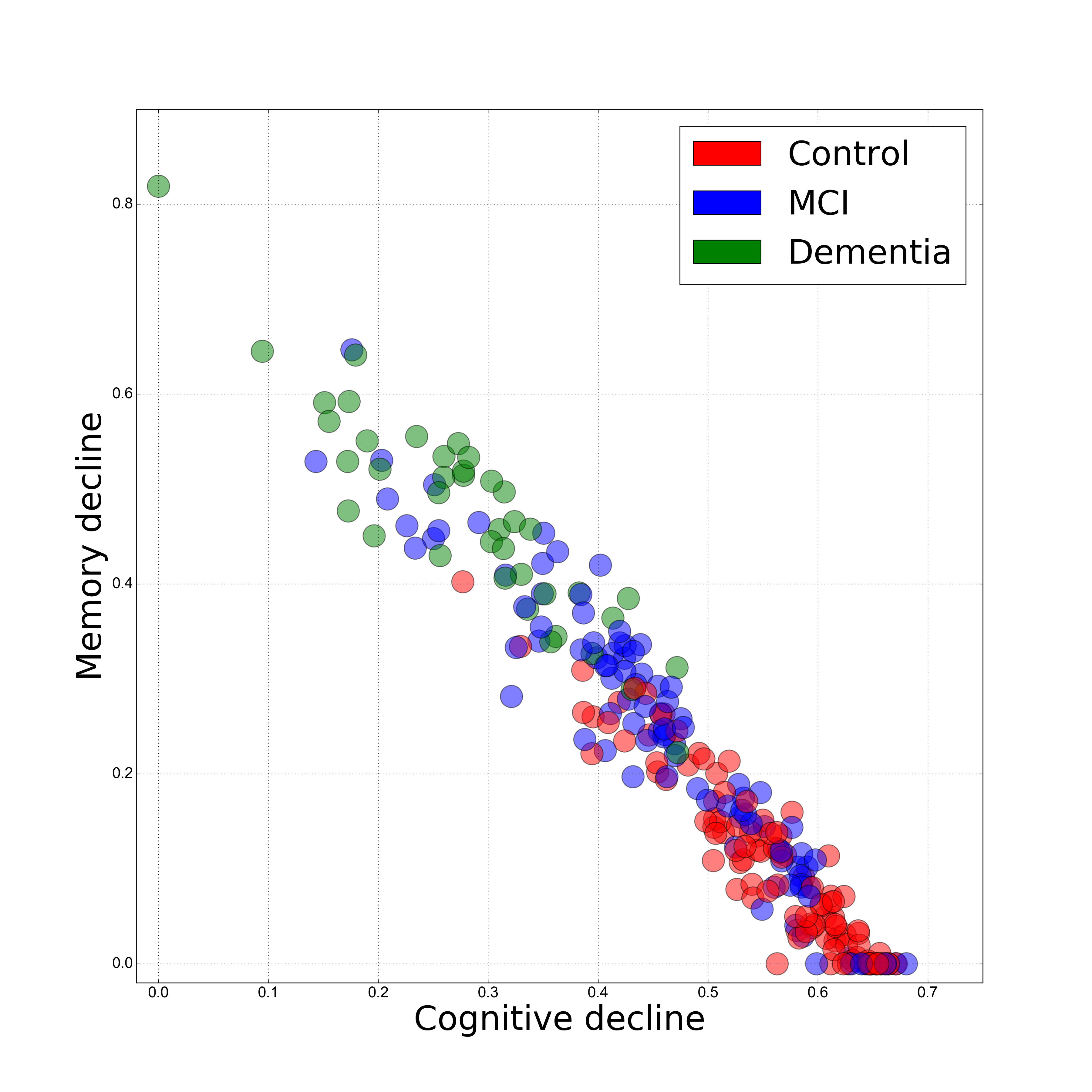}
\end{minipage}
\begin{minipage}{0.52\columnwidth}
    \includegraphics[height = 7cm, width = \columnwidth]{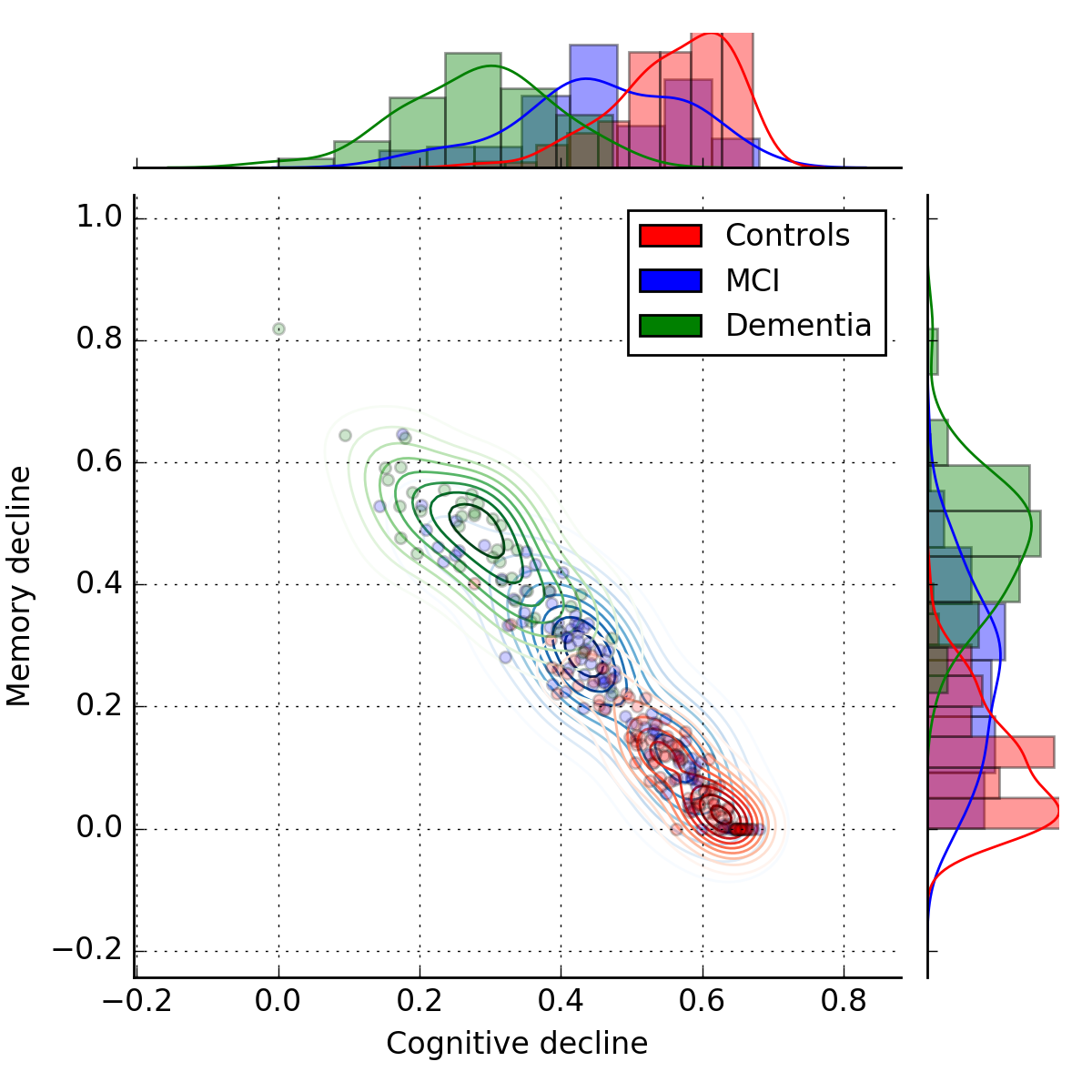}
\end{minipage}
    \caption{\textbf{Left:} Total 248 cases are projected in the Alzheimer’s Disease Progression Space at $48^{th}$  month. Controls are represented in red, MCI in blue and dementia in green. 
   \textbf{Right:} Three different progression rates are identified in MCI and dementia patients at $48^{th}$  month. In this figure low progression rate zone is represented in red, moderate progression rate in blue and high progression rate in green. }
    \label{fig:nmf48}
\end{figure}

\begin{figure}[ht]
    \includegraphics[height = 6cm, width =  15cm]{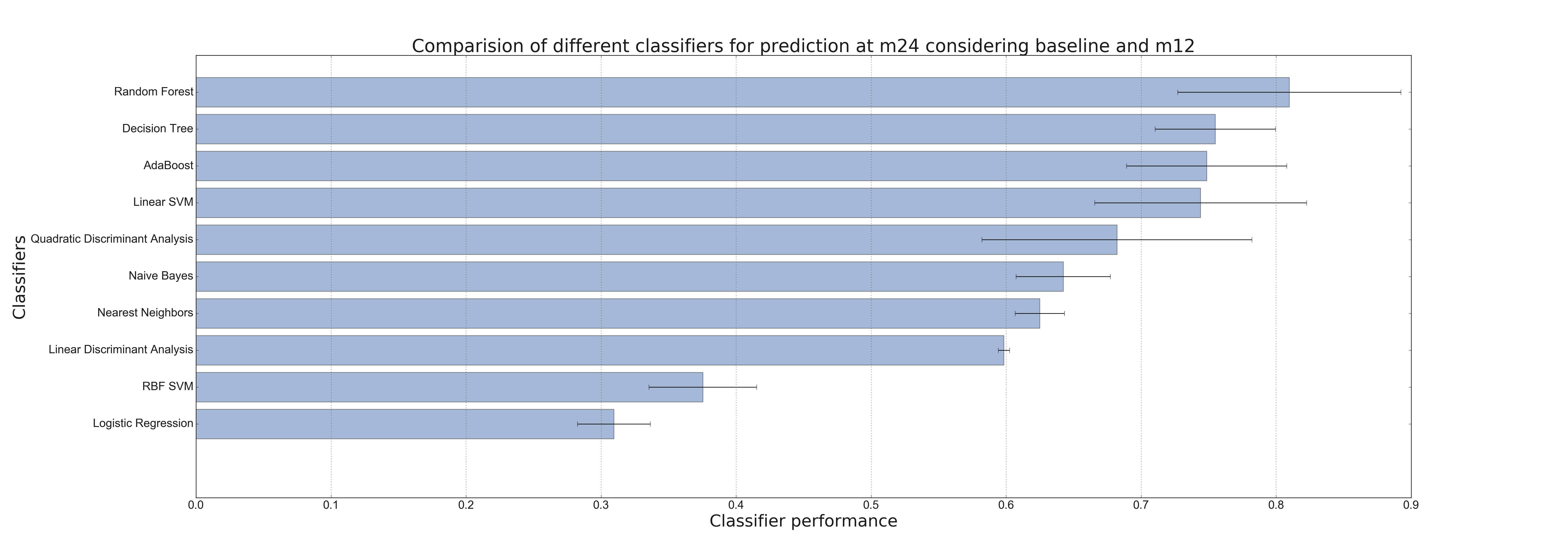}
    \caption{Comparison of different algorithms for the prediction of progression after 24 months from the baseline. Five-fold cross-validation accuracy was used to evaluate each model. Random Forest algorithm provides the highest accuracy. }
    \label{comparison}
\end{figure}

\begin{figure}[ht]
    \includegraphics[height = 6cm, width =  15cm]{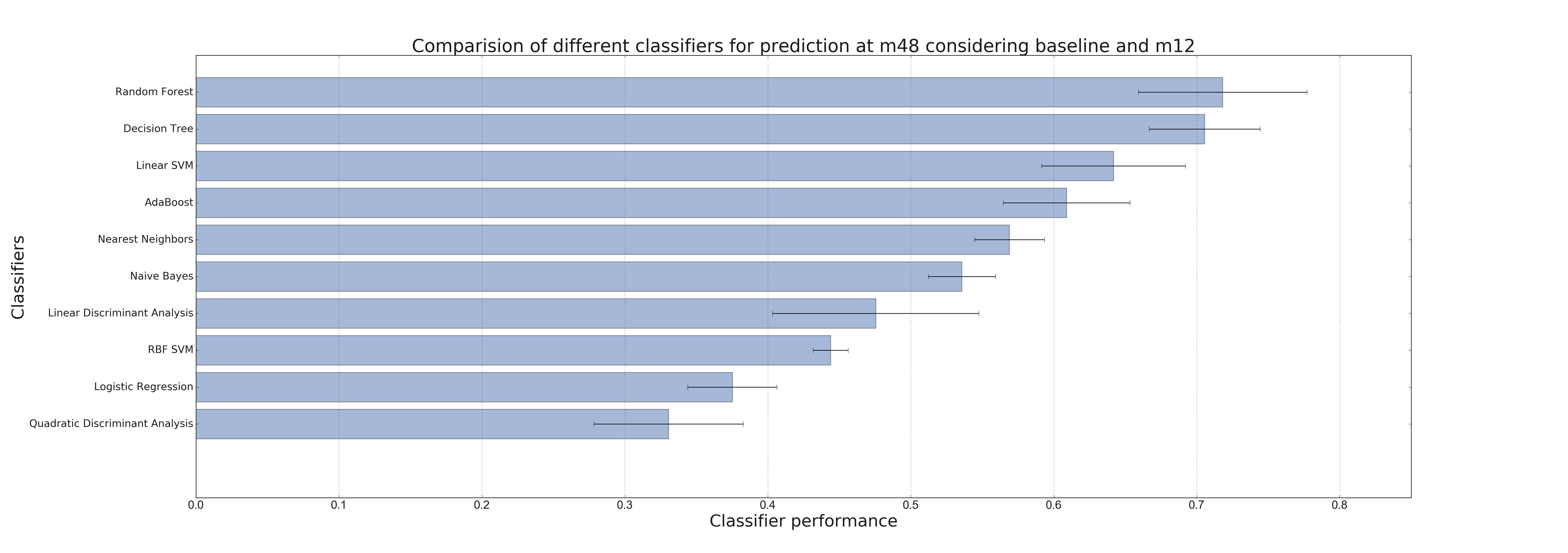}
    \caption{ Comparison of different algorithms for the prediction of progression after 48 months from the baseline. Five-fold cross-validation accuracy was used to evaluate each model. Random Forest algorithm provides the highest accuracy.}
    \label{comparison}
\end{figure}

\begin{figure}[ht]
\centering
\begin{minipage}{0.47\columnwidth}
    \includegraphics[height = 5.6cm, width =  \columnwidth]{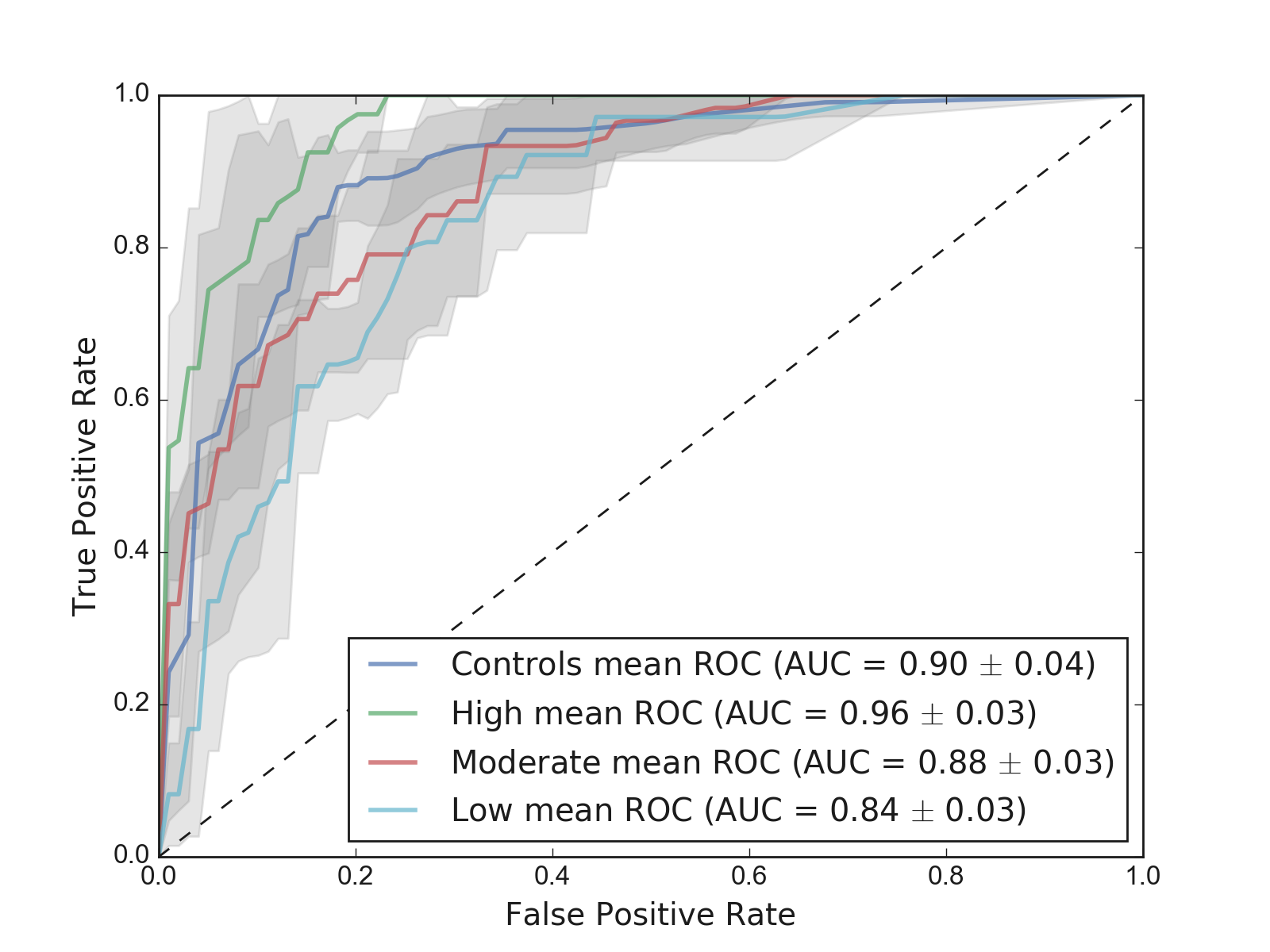}
\end{minipage}
\begin{minipage}{0.40\columnwidth}
    \includegraphics[height = 6.2cm, width = \columnwidth]{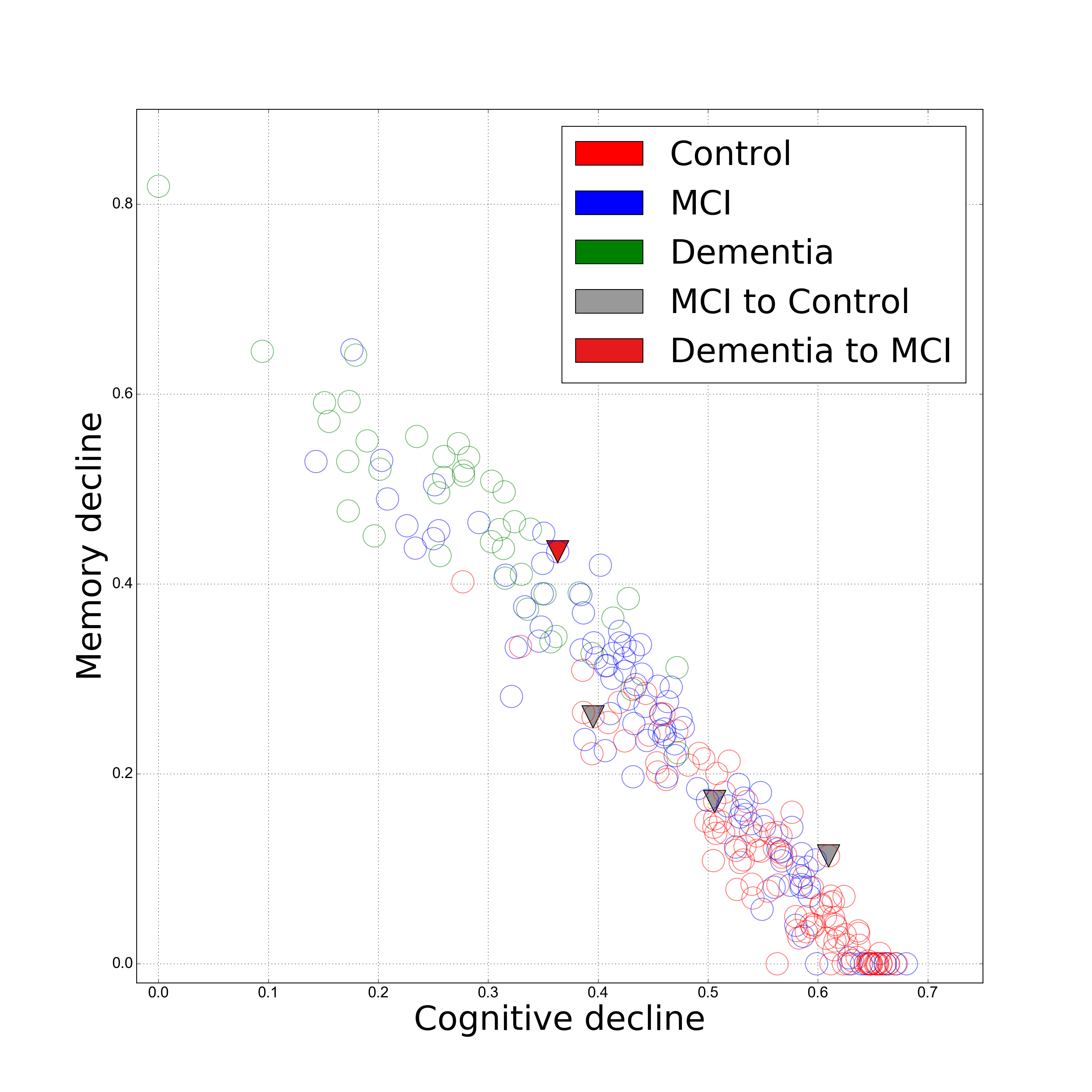}
\end{minipage}
    \caption{\textbf{Left:} Area under ROC curve for each individual class. The predictions for disease stage at $48^{th}$  month were made using Random Forest Algorithm. 
   \textbf{Right:} Shows positions of the reversions (only MCI to control and dementia to MCI) at $48^{th}$  month..}
    \label{fig:nmf}
\end{figure}

\begin{figure}
\centering
\begin{minipage}{0.49\columnwidth}
    \includegraphics[height = 7cm, width =  \columnwidth]{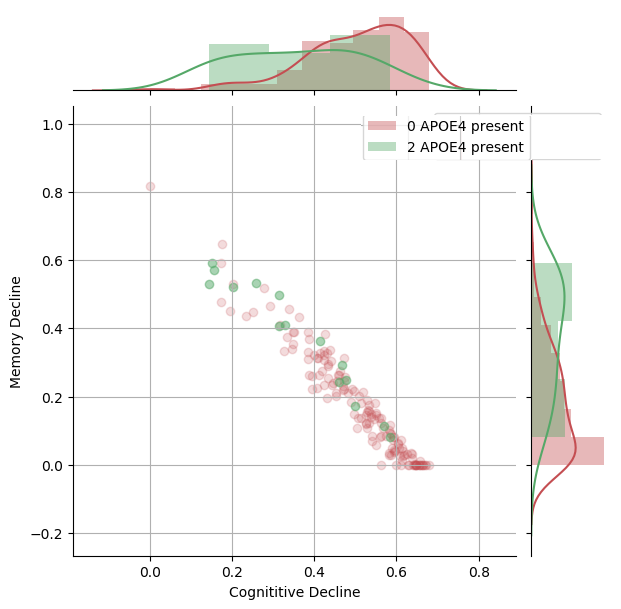}
\end{minipage}
\begin{minipage}{0.45\columnwidth}
    \includegraphics[height = 6.8cm, width = \columnwidth]{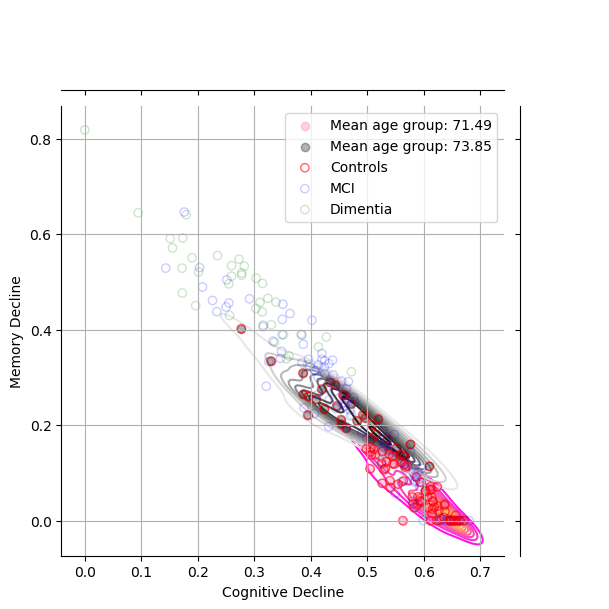}
\end{minipage}
    \caption{\textbf{Left:}  Projection of AD cases with the different number of APOE4 variants to the projection space at $48^{th}$  month.
   \textbf{Right:} Shows the mean age of cases in the two clusters of controls at $48^{th}$  month. Clusters represent aging pattern in controls. The cluster near moderate progression rate zone has high mean age than one which is away from it.  }
    \label{fig:nmf}
\end{figure}

\end{document}